\documentclass{article}
\usepackage{ijcai16}

\usepackage{times}
\usepackage{helvet}
\usepackage{courier}
\usepackage{times}
\usepackage{helvet}
\usepackage{courier}
\usepackage{graphicx}
\usepackage{subfigure}
\usepackage{multirow}%
\usepackage{amsmath,amssymb}
\usepackage{color}
\usepackage{mathrsfs}
\usepackage{amsthm}
\usepackage{eucal}
\usepackage{algorithm}
\usepackage{algorithmic}
 \usepackage{epstopdf}
\def\bzero{{\mathbf 0}}

\def\bb{{\mathbf b}}
\def\bx{{\mathbf x}}

\def\bw{{\mathbf w}}

\def\bg{{\mathbf g}}

\def\br{{\mathbf r}}

\def\bB{{\mathbf B}}

\def\bG{{\mathbf G}}
\def\bX{{\mathbf X}}

\def\bI{{\mathbf I}}

\def\bK{{\mathbf K}}
\def\bM{{\mathbf M}}

\def\bE{{\mathbf E}}

\def\bP{{\mathbf P}}
\def\bQ{{\mathbf Q}}
\def\bR{{\mathbf R}}
\def\bL{{\mathbf L}}

\def\bS{{\mathbf S}}

\def\mR{{\mathbb R}}
\def\mB{{\mathcal B}}

\title{Transfer Hashing with Privileged Information}
\author{Joey Tianyi Zhou$^\dag$, Xinxing Xu$^\dag$, Sinno Jialin Pan$^\ddag$, Ivor W. Tsang$^\S$, \\
{\bf Zheng Qin$^\dag$, and Rick Siow Mong Goh$^\dag$} \\
$^\dag$Institute of High Performance Computing, Singapore \\
$^\ddag$Nanyang Technological University, Singapore \\
$^\S$ QCIS, University of Technology Sydney, Australia \\
{$^\dag$\{zhouty,xuxx,qinz,gohsm\}@ihpc.a-star.edu.sg, $^\ddag$sinnopan@ntu.edu.sg, $^\S$ivor.tsang@uts.edu.au}}
\begin{document}

\maketitle

\begin{abstract}
Most existing learning to hash methods assume that there are sufficient data, either labeled or unlabeled, on the domain of interest (i.e., the target domain) for training. However, this assumption cannot be satisfied in some real-world applications. To address this data sparsity issue in hashing, inspired by transfer learning, we propose a new framework named Transfer Hashing with Privileged Information (THPI). Specifically, we extend the standard learning to hash method, Iterative Quantization (ITQ), in a transfer learning manner, namely ITQ+. In ITQ+, a new slack function is learned from auxiliary data to approximate the quantization error in ITQ. We developed an alternating optimization approach to solve the resultant optimization problem for ITQ+. We further extend ITQ+ to LapITQ+ by utilizing the geometry structure among the auxiliary data for learning more precise binary codes in the target domain. Extensive experiments on several benchmark datasets verify the effectiveness of our proposed approaches through comparisons with several state-of-the-art baselines.
\end{abstract}

\section{Introduction}

Hashing methods have been applied for efficient similarity search in many application areas, especially for information retrieval~\cite{HashSurvey1}. The goal of hashing is to design or learn a compact binary code, with each bit taking values of either -1/1 or 0/1, in a low-dimensional space for each data instance such that similar instances in the original space are mapped to similar binary codes. As a result, data instances can be stored in a low cost, and the similarity between instances can be efficiently computed with the Hamming distance using binary operation (XOR).

Most existing learning to hash methods require a large amount of data instances to learn a set of hash functions to construct binary codes~\cite{HashSurvey1}. However in some real-world applications, for a domain of interest, i.e., the target domain, the data instances may not be sufficient enough to learn a precise hashing model. For example, Taobao.com provides a platform for small businesses and individual entrepreneurs to open online stores. Suppose an individual entrepreneur wants to build a hashing system for all the images of the products sold in his/her online store. Unfortunately, the number of images are not sufficient enough to build a precise hashing system. A straightforward solution is to download images of the same or related products from other e-commerce websites, such as Amazon.com or eBay.com, to help learning a hashing system. However, in many individual online stores in Taobao.com, the images used to demonstrate the products are usually amateur, which are taken by the individual entrepreneur himself/herself, while the images of products demonstrated in Amazon.com are professional. A simple aggregation of these two different kinds of images may not help to learn a precise hashing system for the target images. Motivated by transfer learning~\cite{PanY09TKDE}, instead of using the images downloaded from Amazon.com or other e-commerce websites directly, one can extract knowledge from the auxiliary images, and then transfer the knowledge to help learning a more precise hashing system for the target images.

Specifically, we propose a novel framework named ``Transfer Hashing with Privileged Information'' (THPI),  which is a marriage of transfer learning and a new learning paradigm, namely Learning Using Privileged Information (LUPI), proposed by ~\cite{vapnik2009new}, where privileged information is assumed to be available for each training instance in the training phrase, and missing in the testing phrase.  Our aim is to construct precise hash codes for a target domain instances, e.g., the images of product in an individual online store, by encoding the privileged information from a source domain, e.g., images on products downloaded from Amazon.com, into a learning to hash model. In this example, the target images can be considered as training instances, while auxiliary images of the same or similar products as the target images can be considered as their privileged information in the training phrase. As shown in~\cite{vapnik2015learning}, the amount of training data required for training can be dramatically reduced with the help of privileged information. Therefore, we expect that with privileged information, we are able to learn precise hashing functions for the target domain where data instances are not sufficient.

Note that the proposed THPI framework is different from cross-modal hashing which assumes that training instances of different modalities are rich, and are sufficient to learn reliable hash codes, respectively~\cite{DBLP:conf/ijcai/KumarU11}. In most cross-modal hashing methods, full correspondences across different modalities are required as input~\cite{DBLP:conf/ijcai/KumarU11,DBLP:conf/ijcai/WuYZWW15}. Moreover, the goal of cross-modal hashing is to retrieve relevant data across modalities. In THPI, the goal is to learn a set of good hash functions with insufficient training instances in a target domain, and retrieve relevant information in the target domain.

The novelties of our work are summarized as follows,
\begin{itemize}
\item We propose a novel framework named ``Transfer Hashing with Privileged Information'' (THPI) to alleviate data sparsity in a target domain by transferring knowledge from a source domain.
\item A new algorithm named ITQ+ is proposed, where a novel slack function for incorporating privileged information is introduced to regularize the learning of hash codes for the target domain.
\item We further extend ITQ+ to LapITQ+, where underlying graph structure extracted from the source domain is encoded as a prior for learning more precise hash codes for the target domain.
\end{itemize}

\section{Related Work}
\subsection{Learning to Hash}
Most hashing methods focus on how to quantize data with minimal information loss. For example, Locality-sensitive Hashing (LSH)~\cite{raginsky2009locality} uses a set of random projections followed by thresholding. Spectral Hashing (SH) formulates the quantization as spectral graph partitioning~\cite{DBLP:conf/nips/WeissTF08}, where the graph geometry on original feature space is preserved. Iterative Quantization (ITQ)~\cite{gong2011iterative} is proposed to refine the initial projections, e.g., Principal Component Analysis (PCA)~\cite{tipping1999probabilistic}, Canonical Correlation Analysis (CCA)~\cite{DBLP:journals/neco/HardoonSS04}, such that quantization error can be reduced.  All these methods require sufficient data for learning hash functions for a domain of interest. Different from these methods, THPI aims at alleviating data sparsity and further improving hashing performance on the target domain by exploiting knowledge from other domains, which may be of heterogeneous features. Recently, cross-modal hashing~\cite{DBLP:conf/ijcai/KumarU11,zhang2011composite} that aims to learn hash codes with data from different modalities draws much attention. However, cross-modal hashing assumes training instances of different modalities be sufficient to learn reliable hashing functions respectively, which is different from THPI.

\subsection{Transfer Learning}
Transfer learning (TL)~\cite{PanY09TKDE} aims to transfer knowledge across different domains so that rich source domain knowlege can be used to build better classifiers on a target domain where the transferred knowledge can be labels~\cite{DBLP:conf/aistats/ZhouT14}, features~\cite{sinno11tnn}, cross-domain correspondences~\cite{DBLP:conf/aaai/ZhouPTY14,DBLP:conf/aaai/ZhouPTY16}. TL has shown promising results in many machine learning tasks, such as classification and regression. To the best of our knowledge, there is only one work on studying transfer learning for hashing~\cite{DBLP:conf/mm/OuYLLL14}. Different from their work, we focus on how to transfer knowledge across heterogeneous feature spaces in an unsupervised manner.

\subsection{Learning using Privileged Information}
Recently, \citeauthor{vapnik2009new}~\shortcite{vapnik2009new} introduced a new learning paradigm namely learning using privileged information (LUPI). In LUPI, auxiliary privileged features are assumed to be available in the training phrase but not available in the testing phrase. A new model SVM+ is proposed by exploiting the privileged features to construct a correcting function in traditional Support Vector Machines (SVMs) to control the loss such that the learned classifiers can embrace stronger generalization ability. Specifically, given a set of training data $\{\bx_i,\tilde{\bx}_i\}_{i=1}^{l}$,  where $\tilde{\bx}$ is the corresponding privileged features for the original features $\bx$, SVM+ aims to learn a target classifier $f(\bx) = \bw^\top \bx$ from the original feature vectors and a slack approximation function $\epsilon(\tilde{\bx})=\tilde{\bw}^\top\tilde{\bx}$   from the privileged feature vectors, simultaneously. The goal of the slack approximation function is to control the loss of the target classifier by incorporating privileged features. The objective function of SVM+ is written as follows:
\begin{eqnarray}
\min_{\bw,\tilde{\bw},b} &&\frac{1}{2} (\|\bw\|^2 + \lambda\|\tilde{\bw}\|^2) + C\sum_{i=1}^l\epsilon(\tilde{\bx})\nonumber\\
\mbox{s.t.}&& y_i\bw^\top\bx_i \geq 1- \epsilon(\tilde{\bx_i}), \vee i = 1,2,\cdots,l,\nonumber\\
&&\epsilon(\tilde{\bx}) \geq 0, \vee i = 1,2,\cdots,l,\nonumber
\end{eqnarray}
Inspired by the formulation of SVM+ and recent advances on LUPI~\cite{MDAHSPlus,ITML+,DBLP:conf/iccv/SharmanskaQL13}, in this work, we  aim to construct a slack function for hashing by incorporating privileged information to learn a more precise hash model in the target domain where data is sparse.

\section{Iterative Quantization with Privileged Information (ITQ+)}

Suppose that we are given $n$ data points $\{\bx_{T_1},\bx_{T_2},\ldots,\bx_{T_{n}}\}$ with $\bx_{T_i}\!\in\! \mR^{d_T}$. Denote by $\bX_T = [\bx_{T_1},\ldots,\bx_{T_{n}}] ^\top \!\in\! \mR^{ n\times d_T}$ the data matrix. Without loss in generality, we assume that all the $n$ points have been zero-centered, \emph{i.e.}, $\sum_{i=1}^{n}\bx_{T_i} = \bzero$. The goal of learning to hash is to learn a binary code matrix $\bB_T \in \mR^{n \times c}$ with its elements in $\{-1,1\}$, where $c$ is the length of each hash code. For each bit $k\in\{1,...,c\}$, a binary function $h_{k}(\bx_T) = \text{sgn}(\br_{k}^\top\bx_T)$ is learned, where $\br_{k}\in \mR^{d_T}$ is the hyperplane for the $k$-th bit. Denote by $\bR \in \mR^{d_T\times c}$ the projection matrix for all the $c$ bits, the binary code matrix $\bB_T$ can be obtained by setting $\bB_T = \text{sgn}(\bX_T\bR)$.

When $n$ is small, i.e., the available target training data is limited, the hash codes learned by existing methods may not perform well. How to learn a precise hash model from sparse data is a crucial issue for most existing learning-to-hash algorithms. Inspired by the exciting results of LUPI, which prove that the amount of training data can be significantly reduced with privileged information for training a precise predictive model~\cite{lapin2014learning,vapnik2009new,vapnik2015learning}, we propose a new framework for learning to hash, namely Transfer Hashing with Privileged Information (THPI). In THPI, the data sparsity issue on the target domain is alleviated by using privileged information from an auxiliary domain which is referred to as the source domain. Apart from a target feature vector $\bx_T$, in THPI, we assume that corresponding privileged information $\bx_S\!\in\! \mR^{d_S}$ from the source domain is available for training as well, which means that there are $n$ corresponding data pairs $\{(\bx_{S_1},\bx_{T_1}),(\bx_{S_2},\bx_{T_2}),\ldots,(\bx_{S_n},\bx_{T_n})\}$ for training. Furthermore, we denote by $\bX_{SC} \!=\! [\bx_{S_1},\ldots,\bx_{S_{n}}]^\top \!\in\! \mR^{n\times d_S}$ the matrix of the $n$ corresponding instances on the source domain, and $\bX_{SU} \!=\! [\bx_{S_{n+1}},\bx_{S_{n+2}},\ldots,\bx_{S_{n+n_S}}]^\top \!\in\! \mR^{n_S\times d_S}$ the matrix of the additional $n_S$ source domain instances. Note that the privileged information is only available for training but not for testing.

\subsection{Iterative Quantization}
The ITQ algorithm~\cite{gong2011iterative} aims to construct hashing functions using an iterative quantization method to learn a rotation matrix by minimizing the quantization error. Specifically, an orthogonal projection matrix $\bR_T \in \mR^{d_T\times c}$ is learned with the code matrix $\bB_T  \in \{-1,1\}^{n\times c}$ by optimizing the following quantization loss:
\begin{eqnarray}
\min_{\bB_T,\bR_T} && \|\bB_T -  \bX_T \bR_T\|^{2}_{F} \\
\mbox{s.t.} && \bR_T^\top\bR_T = \bI. \nonumber
\end{eqnarray}
%where $\bX \in \mR^{n\times c}$ denotes the target data matrix, $\bB \in [-1,1]^{n\times c}$ and $\bR \in \mR^{c \times c}$.

\subsection{The objective function for ITQ+}
Here, we assume that for $\bB_T$, each bit is balanced, i.e., $\bB_T \in \mB = \{\bB\in\{-1,1\}^{n\times c}\mbox{ and }\sum_{i=1}^{n} \bb_{i} = \bzero\}$, where $\bb_i$ is the $i$-th column of $\bB$. Define $\bE =  \bB_T -  \bX_T \bR_T$, where $\bE$ is the error matrix induced by the quantization process. With the privileged data $\bX_{SC}\in \mR^{n\times c}$, we aim to approximate the quantization error matrix $\bE$ by using a slack function $\bg(\bX_{SC}) = \bX_{SC} \bP$, where $\bP \in \mR^{d_S\times c}$ is another orthogonal projection matrix to be learned. Therefore, we formulate Iterative Quantization with privileged information (ITQ+) as
\begin{eqnarray}\label{eqn:ITQ+}
\min_{\bB_T \in \mathcal{B},\bR_T,\bP_T} \!\!\!&& \frac{1}{2}\|\bE\|^{2}_{F}+ \lambda_1  \|\bE -\bg(\bX_{SC})\|^{2}_{F}  \\
\mbox{s.t.} && \bR_T^\top\bR_T = \bI, \mbox{ and } \bP^\top \bP = \bI \nonumber
\end{eqnarray}
where $\lambda_1 > 0$ is a tradeoff parameter. Note that in SVM+, the privileged information is used to approximate the slack variables, which can be considered as tolerance functions that allow the margin constraints to be violated. Here, in ITQ+, we borrow the high-level idea of SVM+ to use source-domain information to approximate the target-domain quantization error $\bE$.
On one hand, the constructed slack function models the difficulty in quantizing the target domain data with privileged information from the source domain. On the other hand, the constructed slack function can provide a way to regularize the quantization error to avoid overfitting when the size of target domain training data is small.

\subsection{Optimization}
The solution for the optimization problem (\ref{eqn:ITQ+}) can be obtained by alternatingly updating the binary code matrix $\bB_T$ and
the rotation matrices $\bR_T$ and $\bP$. The procedure is summarized in Algorithm~\ref{algo:Sol_ITQ+}, and the details are described in this section.

\subsubsection{Update $\bB_T$ by fixing $\bR_T$ and $\bP$}
By fixing $\bR_T$ and $\bP$, the binary code matrix $\bB_T$ can be obtained by solving the following optimization problem,
\begin{eqnarray}\label{eqn:ITQ+_B}
\min_{\bB_T\in \mB} & & \|\bB_T -  \bX_T \bR_T\|^{2}_{F} \nonumber \\
                    & & + \lambda_1  \|(\bB_T -  \bX_T \bR_T) - \bX_{SC} \bP\|^{2}_{F}.
\end{eqnarray}
As $\bR_T$ and $\bP$ are constant when optimizing $\bB_T$, (\ref{eqn:ITQ+_B}) can be reformulated as
\begin{eqnarray}\label{eqn:ITQ+_sub_B}
\max_{\bB_T\in \mB} \mbox{tr}\left(\bB_T(\lambda_1 \bP^\top \bX_{SC}^\top+ (\lambda_1+1)\bR_T^\top \bX_T^\top)\right),
\end{eqnarray}
where $\mbox{tr}(\cdot)$ denotes the trace of a matrix. The solution for (\ref{eqn:ITQ+_B}) can be obtained by sorting the matrix $\bM$, column-wisely, and then projecting the sorted matrix onto the constraint $\mB$, where
\[\bM=\lambda_1 \bP^\top \bX_{SC}^\top+ (\lambda_1+1)\bR_T^\top \bX_T^\top.\]

\subsubsection{Update $\bR_T$ by fixing $\bP$ and $\bB_T$}
By fixing $\bB_T$ and $\bP$, the optimization problem with respect to $\bR_T$ can be written as follows,
\begin{eqnarray}\label{eqn:sub_Rb}
\min_{\bR_T} &&  \|\bB_T -  \bX_T \bR_T\|^{2}_{F} \nonumber \\&& + \lambda_1  \|(\bB_T -  \bX_T \bR_T) - \bX_{SC} \bP\|^{2}_{F}  \\
\mbox{s.t.} && \bR_T^\top\bR_T = \bI.\nonumber
\end{eqnarray}
As $\bP$ is fixed, and only $\bR_T$ is to be optimized, we further rewrite the above optimization problem as
\begin{eqnarray}
\min_{\bR_T} &&  \left\|\bX_T \bR_T -  \left( \bB_T - \frac{\lambda_1}{\lambda_1+1} (\bX_{SC} \bP) \right)\right\|_{F}^{2} \\
\mbox{s.t.} && \bR_T^\top\bR_T = \bI,\nonumber
\end{eqnarray}
which is an orthogonal procrustes problem~\cite{schonemann1966generalized}, and can be solved analytically. To be specific, by performing the Singular Value Decomposition (SVD) on $\left( \bB_T - \frac{\lambda_1}{\lambda_1+1} (\bX_{SC} \bP) \right)^\top \bX_T$, i.e.,
\[\left( \bB_T - \frac{\lambda_1}{\lambda_1+1} (\bX_{SC} \bP) \right)^\top \bX_T =  \hat{\bS}\Sigma \bS^\top,\]
we obtain
\begin{eqnarray}\label{eqn:sol_Rb}
\bR_T = \hat{\bS} \bS^\top.
\end{eqnarray}
\subsubsection{Update $\bP$ by fixing $\bR_T$ and $\bB_T$}
With $\bR_T$ and $\bB_T$ fixed, we obtain an optimization problem with respect to $\bP$ as
\begin{eqnarray}\label{eqn:sub_P}
\min_{\bP}&& \|(\bB_T -\bX_T \bR_T) - \bX_{SC} \bP\|^{2}_{F}\\
\mbox{s.t.} && \bP^\top\bP = \bI,\nonumber
\end{eqnarray}
which again is a standard orthogonal procrustes problem, and can be solved analytically as follows,
\begin{eqnarray}\label{eqn:sol_P}
\bP = \hat{\bQ} \bQ^\top,
\end{eqnarray}
where $\left(\bB_T -\bX_T \bR_T \right)^\top \bX_{SC}  = \hat{\bQ} \Lambda \bQ^\top$ is obtained by performing the SVD on $\left(\bB_T -\bX_T \bR_T \right)^\top \bX_{SC}$.

\begin{algorithm}[tbp]
\caption{: Alternating optimization procedure for ITQ+ (or LapITQ+)}
\begin{algorithmic}[1]
\STATE Initialize $\bR_T^{0},\bP^{0}$ to be the random orthogonal matrices, and set $\tau = 0$.
\WHILE{not converge}
\STATE Update $\bB_T^{\tau+1}$ by solving~(\ref{eqn:ITQ+_sub_B}) (or (\ref{eqn:sub_B:lap}) for LapITQ+).
\STATE Update $\bR_T^{\tau+1}$ according to (\ref{eqn:sol_Rb}).
\STATE Update $\bP^{\tau+1}$ according to (\ref{eqn:sol_P}).
\STATE $\tau = \tau + 1$.
\ENDWHILE
\end{algorithmic}
\label{algo:Sol_ITQ+}
\end{algorithm}

\section{Extension for ITQ+ (LapITQ+)}
In ITQ+, only $\bX_{SC}$ is used for learning a hashing model for the target domain. In practice, besides $\bX_{SC}$, we may have a large amount of training instances $\bX_{SU}$ on the source domain, whose corresponding feature vectors in the target domain are unknown. To fully exploit all the source domain data to learn a more precise hashing model for the target domain, we proposed an extension of ITQ+ in this section, namely LapITQ+.

Our motivation is from multi-view learning, where the underlying graph structures in different views are assumed to be similar~\cite{he2011graph}. Intuitively, as we have a large amount of training instances on the source domain, we can learn a precise graph structure for the source domain, and encode the structure as a regularization term for learning the hash codes for the target domain. Specifically, we can first apply ITQ to learn the hash codes $\bB_S$ using all the available source domain data $\bX_S = [\bX_{SC}^\top,\bX_{SU}^\top]^\top$ by optimizing the following quantization loss,
\begin{eqnarray}
\min_{\bB_S,\bR_S} && \|\bB_S -  \bX_S \bR_S\|^{2}_{F} \nonumber\\
\mbox{s.t.} && \bR_S^\top\bR_S = \bI,
\end{eqnarray}
where $\bX_S \in \mR^{(n_S+n)\times d_S}$, $\bB_S \in \{-1,1\}^{(n_S+n)\times c}$ and $\bR_S \in \mR^{d_S \times c}$. This can be done offline in advance.

Next, we construct an adjacency graph $\bG$ from the hash codes $\bB_S$ as follow: for each code $\bB_{S_{i}}$ (each row of $\bB_S$), connect $\bB_{S_{i}}$ to its $k$ nearest neighbors with a weight of value $1$, where Hamming distance is applied. After constructing the adjacency graph, we  use it to define the graph Laplacian $\bL_C \in \mR ^{n \times n}$ for the target domain data. Finally, the proposed LapITQ+ method is formulated as
\begin{eqnarray}\label{eqn:LapITQ+}
&
\min\limits_{\substack{
    \bB_T \in \mathcal{B}, \\
    \bR_T,\bP_T}}
& \|\bE\|^{2}_{F}+ \lambda_1  \|\bE -\bg(\bX_{SC})\|^{2}_{F} + \lambda_2  tr(\bB_T^\top\bL_C\bB_T) \nonumber \\
& \mbox{s.t.} & \bR_T^\top\bR_T = \bI, \mbox{ and } \bP^\top \bP = \bI. \nonumber
\end{eqnarray}
where $\lambda_1,\lambda_2\geq 0$ are parameters. Compared to (\ref{eqn:ITQ+}), the third term in the above objective is to transfer the graph structure from the source domain to the target domain. Note that instead of constructing the graph Laplacian matrix on the original space~\cite{DBLP:conf/nips/WeissTF08}, we construct the graph Laplacian on the binary code space. In this way, we quantify the local properties of the manifold on the source domain and thus naturally transfer the graph structure across domains for learning hash codes. Though learning the hash codes on the source domain and the constructing adjacency graph can be very expensive, they can be done offline. The procedure to solve the above optimization is the same as that used for ITQ+ except for the update on $B_T$.

\subsection{Updating $B_T$ by fixing $\bP$ and $\bR_T$}
We first relax the constraint $\bB_T \in \{-1,1\}^{n\times c}$ to $\bB_T\in [-1,1]^{n\times c}$ on the feasible domain $\mB$, and obtain the following constrained quadratic programming optimization,
\begin{eqnarray}\label{eqn:sub_B:lap}
\min_{\bB_T\in \mB} && -2tr\left(\bB_T \bK \right)+\lambda_2\|\bB_T\bL\|_F^2.
\end{eqnarray}
where $\bK=((1+\lambda_1)\bR^\top\bX_T^\top)+\lambda_1\bP^\top\bX_{SC}^\top$, and $\bL_C = \bL^\top \bL$. Finally, we binarize the codes by
$\bB_T = \mbox{sgn}({\bB}_T)$.

\section{Complexity Analysis}
The computational cost for proposed algorithms mainly depends on two parts: 1) to optimize the binary codes $\bB_T$ and 2) to optimize
the orthogonal rotation matrices $\bR_T$ and $\bP$. For updating $\bB_T$, the time complexity for ITQ+ which involves sorting is $O(n_T log(n_T)c)$, and for LapITQ+, which involves QP programming, the time complexity is $O(n_T^3)$. For updating the orthogonal rotation matrices $\bR_T$ and $\bP$, the time complexities are bounded by $O(c^2 d_T + d_T^3)$ and $O(c^2 d_S + d_S^3)$ respectively. In transfer learning, $n_T$ is supposed to be not large. Moreover, in learning to hash, the code length $c$ is supposed to be small. The dimensions $d_S$ and $d_T$ can be preprocessed to be small through dimensionality reduction techniques, such as CCA or PCA. Therefore, the overall complexities for ITQ+ and LapITQ+ are reasonably small.

\section{Experiments}
\subsection{Datasets and Experimental Setup}
To verify the  effectiveness of our proposed approaches, ITQ+ and LapITQ+, we conduct a series of experiments on three benchmark datasets: BBC Collection~\cite{greene06icml}, multilingual Reuters~\cite{DBLP:conf/nips/AminiUG09}, and  NUS-WIDE~\cite{nus-wide-civr09}.

\textbf{BBC Collection} was collected for multi-view learning, where each instance is represented by three views. Specifically, this dataset was constructed from a single-view BBC corpora by splitting news article into related ``views'' of text. %The original data are pre-processed with stemming, stop-word removal and low term frequency filtering (count $< 3$).
On this dataset, we consider View 1 as the source domain, and View 2 as the target domain.

\textbf{Multilingual Reuters Collection} is a text dataset with over 11,000 news articles from 6 categories in 5 languages, e.g., English, French, etc., which are represented by a bag-of-words weighted by TF-IDF. Each document was also translated into the other four languages to construct correspondences. In the experiments, we  use the English as the source domain and French as the target domain. Note that the original data is of very high dimensionality,  we first perform PCA with $60\%$ energy preserved on the TF-IDF features. After that, we obtain 1131- and 1230-dimensional features for the the English and French documents respectively.

\textbf{NUS-WIDE dataset} consists of 269,648 images from 81 concepts with a total number of 5,018 unique tags downloaded from Flickr. Following~\cite{song2013inter}, we use 150-D color moment for each image. For the corresponding text documents, we use bag-of-word features based on the 5,018 tags provided by NUS-WIDE, and further reduce its dimensionality by LDA to obtain a 60-D textual feature vector for each document. On this dataset, we treat image features as the target data and the text features as the source data.

While different features may lead to different retrieval performances, the evaluation of different features is not the focus  of this paper. To simulate the partial cross-domain correspondence setting, we randomly
select a fraction of training examples that are of both modalities as correspondences and denote the correspondences ratio by $\alpha$, i.e., $\alpha = n/(n+n_S)$, and from the remaining data, we randomly selected 10\% as the test samples. The parameters $\lambda_1$ and $\lambda_2$ for the proposed methods are tuned by cross validation in the range of $[0,0.001,0.005,\cdots, 1,2]$.
We set the maximum number of iterations to be 150. To remove any randomness caused by random selection of training set, the results are averaged over 10 training-testing splits.
To assess the performance of different algorithms, following the evaluation protocols in  \cite{gong2011iterative,raginsky2009locality}, a nominal threshold of the average distance to the 50th nearest neighbor is used to determine whether a database point returned for a given query is considered a true positive. Finally, we adopt the widely used criterion Mean Average Precision (MAP) \cite{gong2011iterative,DBLP:conf/ijcai/KumarU11} for evaluation.

\begin{figure*}[t!]
	\vspace{-5mm}
\centering
\subfigure[BBC] {\label{fig:BBC}\includegraphics[width=0.3\textwidth]{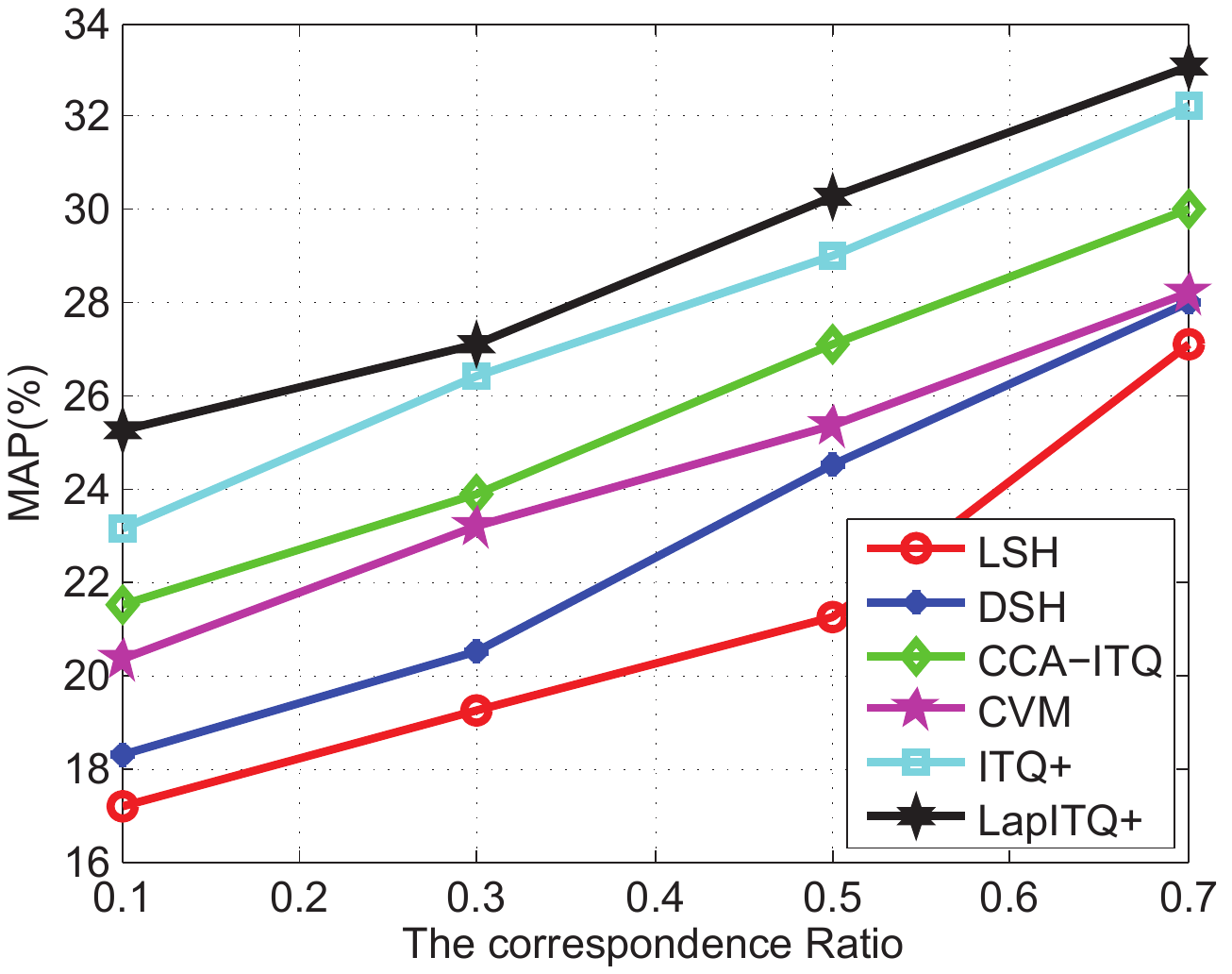}}\hspace{3mm}
\subfigure[Reuters] {\label{fig:Reuters}\includegraphics[width=0.3\textwidth]{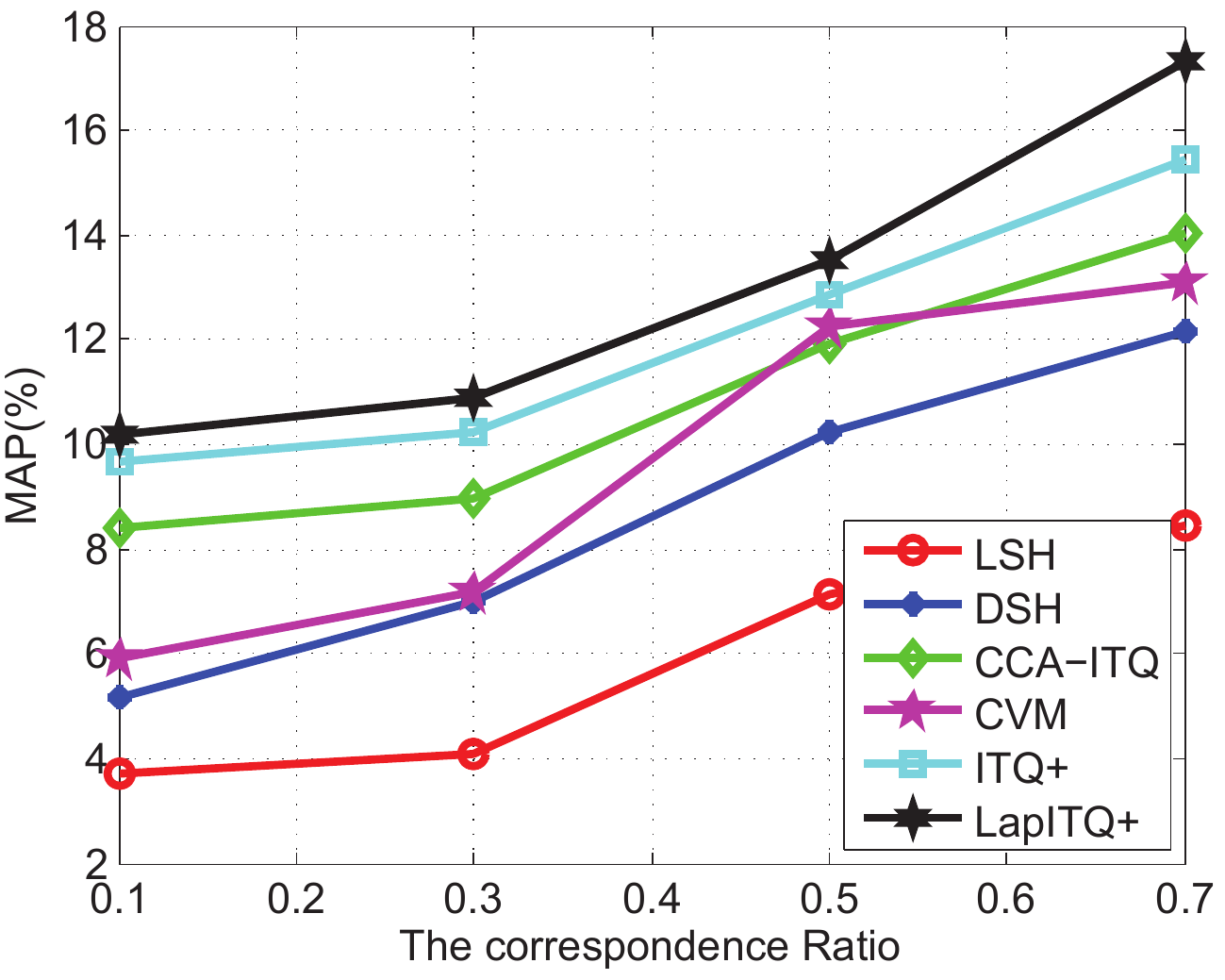}}\hspace{3mm}
\subfigure[NUS-Wide] {\label{fig:NUS}\includegraphics[width=0.3\textwidth]{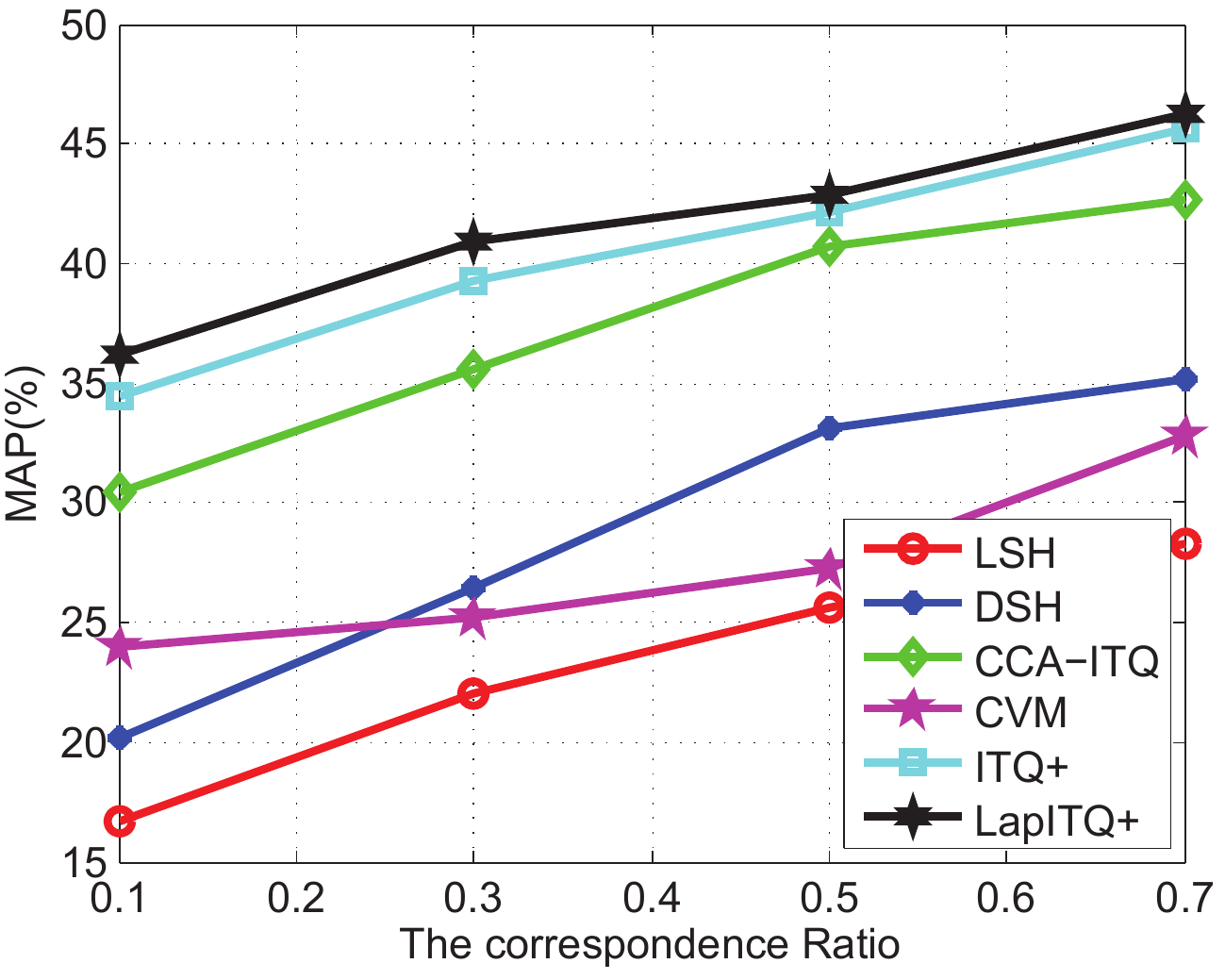}}\centering
\caption{MAP v.s. Privileged Data Size.}\label{fig:Correspondence}
\end{figure*}

\begin{figure*}[t!]
\centering
\subfigure[BBC] {\label{fig:BBC500}\includegraphics[width=0.3\textwidth]{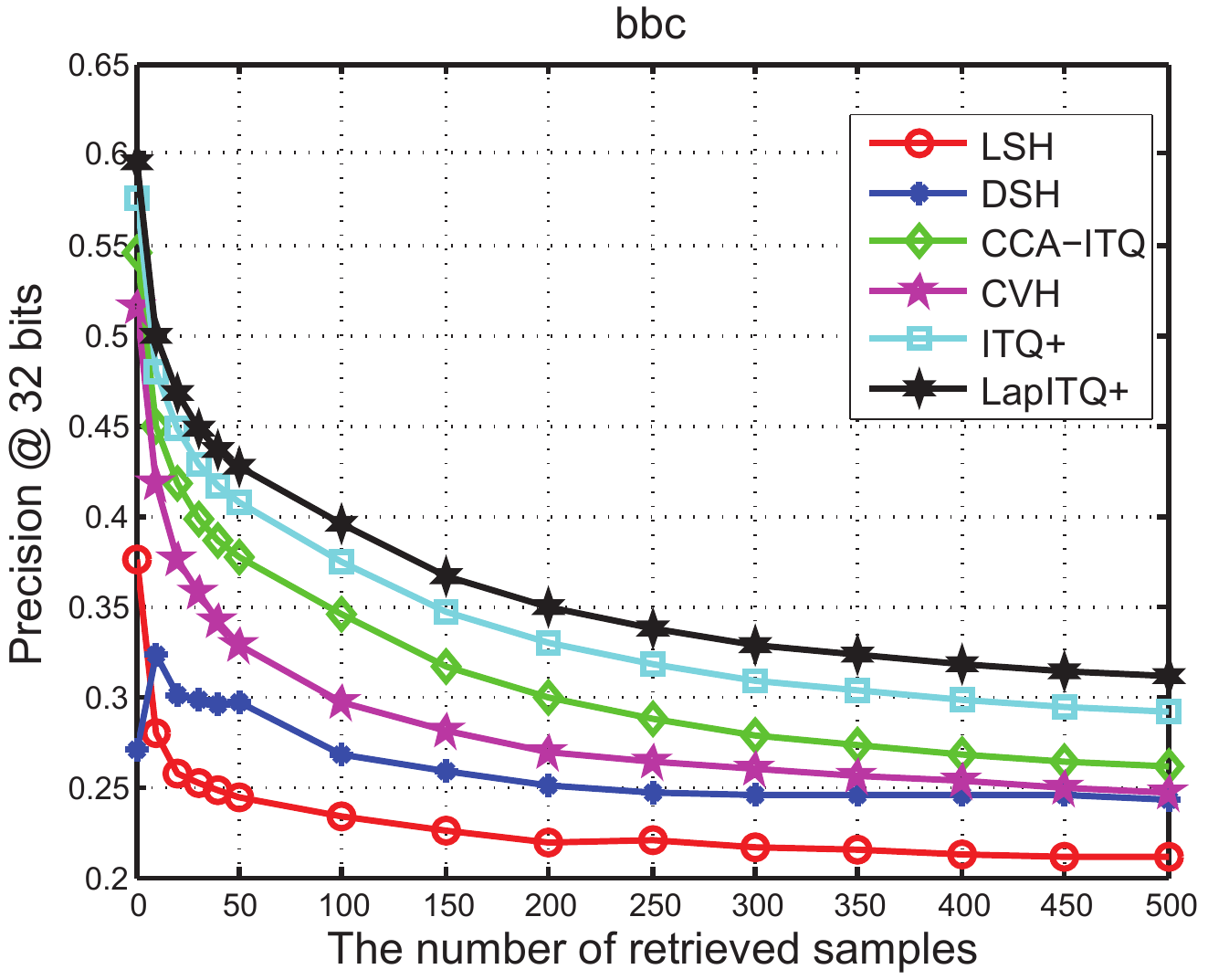}}\hspace{3mm}
\subfigure[Reuters] {\label{fig:Reuters500}\includegraphics[width=0.3\textwidth]{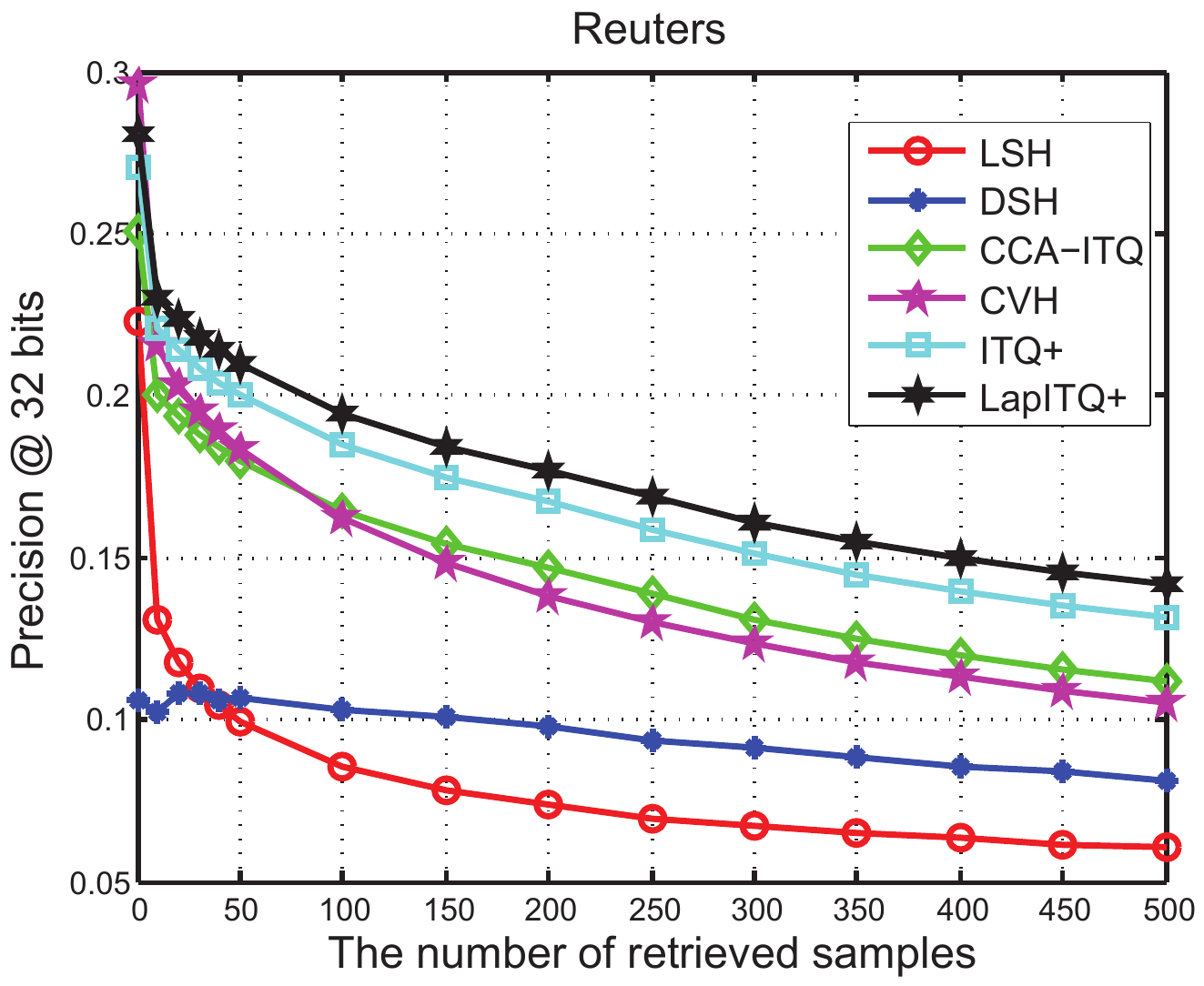}}\hspace{3mm}
\subfigure[NUS-Wide] {\label{fig:NUS500}\includegraphics[width=0.3\textwidth]{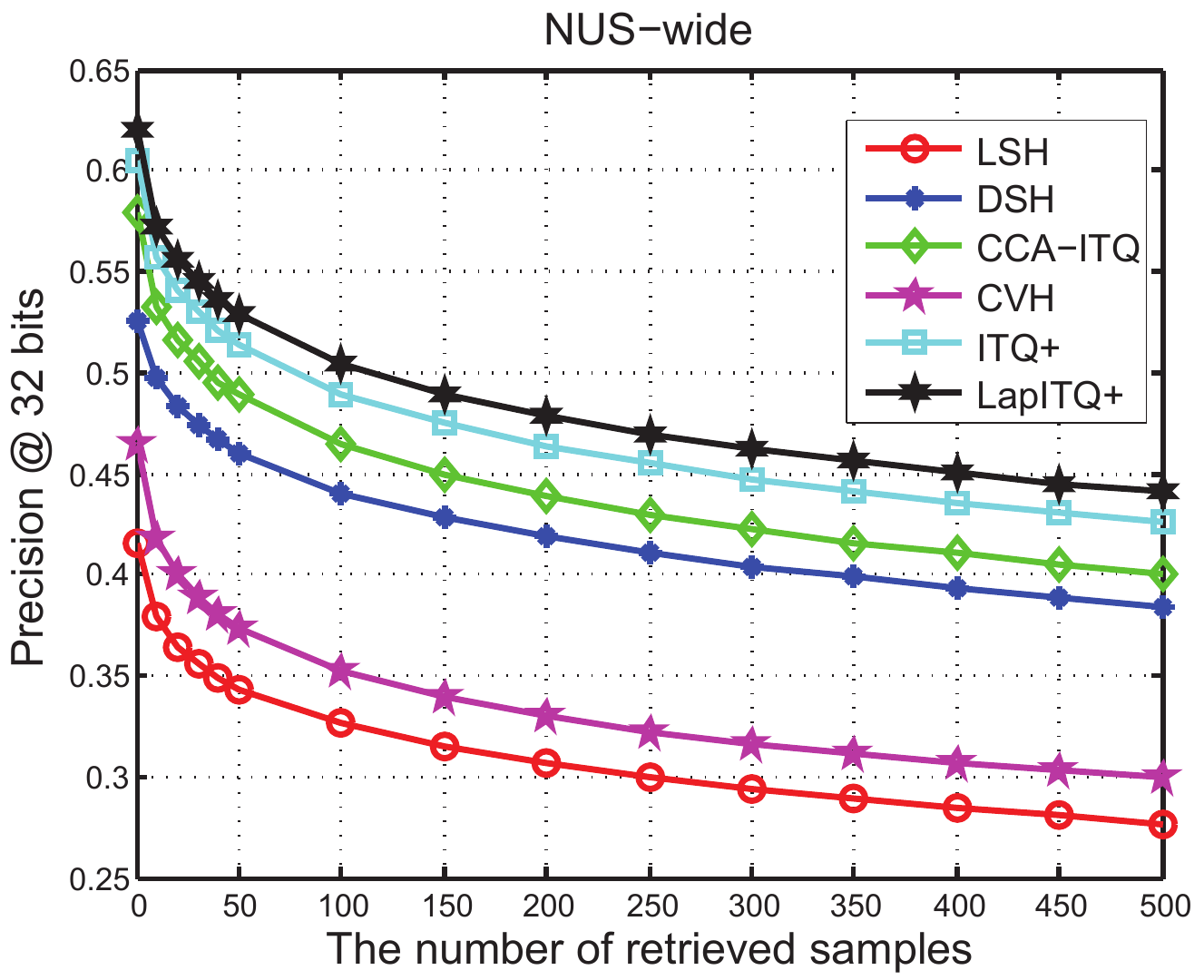}}
\caption{Precision v.s. The Number of Retrieved Samples.}\label{fig:retr500}
\end{figure*}
\subsection{Compared Methods and Evaluation}
We first evaluate the performance of different methods
by varying the number of hashing bits in the range of
$\{8, 16, 32, 64\}$, with fixed $\alpha = 0.5$. The proposed transfer hashing approach is compared with four state-of-the-art hashing methods, i.e., LSH \cite{andoni2006near}, DSH \cite{jin2014density}, CCA-ITQ \cite{gong2011iterative}  and one cross-modal hashing method   CVH \cite{DBLP:conf/ijcai/KumarU11}.
\begin{itemize}
\item \textbf{LSH:} Local Sensitive Hashing (LSH)  \cite{andoni2006near} is based on the a series of random projection to preserve pairwise distances for data points.
\item \textbf{DSH:} Density Sensitive Hashing (DSH) \cite{jin2014density}  is to exploit the clustering results to generate a set of candidate hash functions and to select the hash functions which can split the data most equally.
\item \textbf{CCA-ITQ:}  For a fair comparison, we utilize the data from two domains by using CCA instead of PCA to generate initializations for ITQ~\cite{gong2011iterative}.
\item \textbf{CVH:} Cross-View Hashing (CVH) \cite{DBLP:conf/ijcai/KumarU11} extends Spectral Hashing \cite{DBLP:conf/nips/WeissTF08} in a cross-modal manner, which maps similar data to similar codes across views based on a similarity graph.
\end{itemize}

We report MAP over all the test data for different methods  in  the Table \ref{tab:results}. From the results, we can see that the proposed methods ITQ+ and LapITQ+ perform much better than the baselines. The reason is that ITQ+ and LapITQ+ introduce the new slack function to regularize the quantization loss by using privileged information, which is very important for loss generalization, and insensitive to noise or outliers, especially when the target data is sparse. Compared to LSH and DSH, both ITQ and CVH show better performance as they can make use of knowledge of both the source domain and the target domain by projecting them onto a common space. In this way, the source domain data can be utilized to slightly alleviate the data sparsity issue on the target domain. However these two methods still show inferior performance compared to ITQ+ and LapITQ+. As CCA-ITQ simply performs CCA as a preprocessing step, CCA does not explicitly affect the quantization loss during learning of hashing codes. The CVH method extends spectral hashing in a cross-modal manner and learns hashing codes by performing an eigenvalue decomposition, which usually requires a large number of training data.

From the experimental results, we conclude that the proposed slack function is a better way to transfer source domain knowledge for hashing. Most cross-modal hashing methods require a lot of cross-domain data correspondences, and learn hashing functions only on the correspondences. In contrast, LapITQ+ utilizes all source domain data including unparalleled data to learn source-domain hash codes offline, and use the structure underlying these hash codes to regularize the learning of hash codes on the target domain. Finally, we also observe that LapITQ+ outperforms ITQ+ by incorporating data geometry structure, and thus consistently obtain improvement by 1-2$\%$ in MAP.
\begin{table}[!t]
\small
  \centering
  \caption{MAP ($\%$) over 10 runs  with $\alpha = 0.5$.}
       \begin{tabular}{c|c|c|c|c||c|c}
    \hline
   \multicolumn{7}{c}{BBC}\\\hline
    Bit & LSH & DSH & CCA-ITQ & CVH & ITQ+ & LapITQ+ \\
    \hline
    8  & 14.00 & 20.28 & 21.58 & 17.76 & 24.32 & \textbf{26.62}  \\
    16 & 16.79 & 23.25 & 24.69 & 24.61 & 27.69 &\textbf{28.06} \\
    32 & 21.71 & 25.00 & 27.09 & 25.36 & 29.00 & \textbf{30.30}\\
    64 & 24.61 & 28.36 & 26.51 & 27.62 & 29.58 & \textbf{31.33}\\
    \hline
   \multicolumn{7}{c}{Reuters}\\\hline
    Bit & LSH & DSH & CCA-ITQ & CVH & ITQ+ & LapITQ+ \\
    \hline
    8 &  6.37 & 7.10 & 8.75  & 8.46 & 9.67 & \textbf{10.12}  \\
    16 & 6.55 & 8.12 & 10.00 &9.49 & 10.96 &\textbf{11.32} \\
    32 & 7.17 & 8.15 & 11.93 & 9.91 & 12.85 & \textbf{13.51}\\
    64 & 7.24 & 7.81 & 13.06 & 12.97 & 13.95 & \textbf{14.70}\\
    \hline
       \multicolumn{7}{c}{NUS-wide}\\\hline
    Bit & LSH & DSH & CCA-ITQ & CVH & ITQ+ & LapITQ+ \\
    \hline
    8 & 14.86 &  23.82 & 33.18 & 21.15 & 35.07 & \textbf{35.42}  \\
    16 & 20.49 & 25.69 & 36.99 & 24.49 & 38.51 & \textbf{39.13} \\
    32 & 25.58 & 33.10 & 40.68 & 27.30 & 42.16 & \textbf{42.85}\\
    64 & 28.50 & 35.42 & 42.45 & 30.57 & 44.01 & \textbf{45.40}\\
    \hline
    \end{tabular}\label{tab:results}
\end{table}

\subsection{Training Data Size and Retrieved Sample Size}
We randomly select  $[10\%,30\%,50\%,70\%]$ of
the data from the target domain  as the training set to evaluate the influence of
training size on all the methods. Furthermore, for these data, we are also given the corresponding privileged data during training. Correspondences
ratio is set to be $\alpha = [0.1,0.3,0.5,0.7]$ accordingly. Results are reported in the Figure \ref{fig:Correspondence}. From the figure, we observe that  ITQ+ and LapITQ+ outperform other baselines by large margins, especially when the target training data size is small.  Although CCA-ITA shows promising results compared with other baselines, it still performs worse than our proposed methods.

In the application of information retrieval, users are usually more interested in precision at the first $K$ returned results. We also report the Top-K precision \cite{liu2011hashing} with varying numbers of $K$ retrieved samples on the three datasets with 32 bits in Figure \ref{fig:retr500}. As can be seen from the figure, our proposed methods achieve the best precisions for different values of $K$. For different numbers of bits, we have similar observations. Thus, we do not report the results here due to space limitation.

\begin{figure}[t!]
\centering
\subfigure[$\lambda_1$] {\label{fig:lam1}\includegraphics[width=0.45\columnwidth]{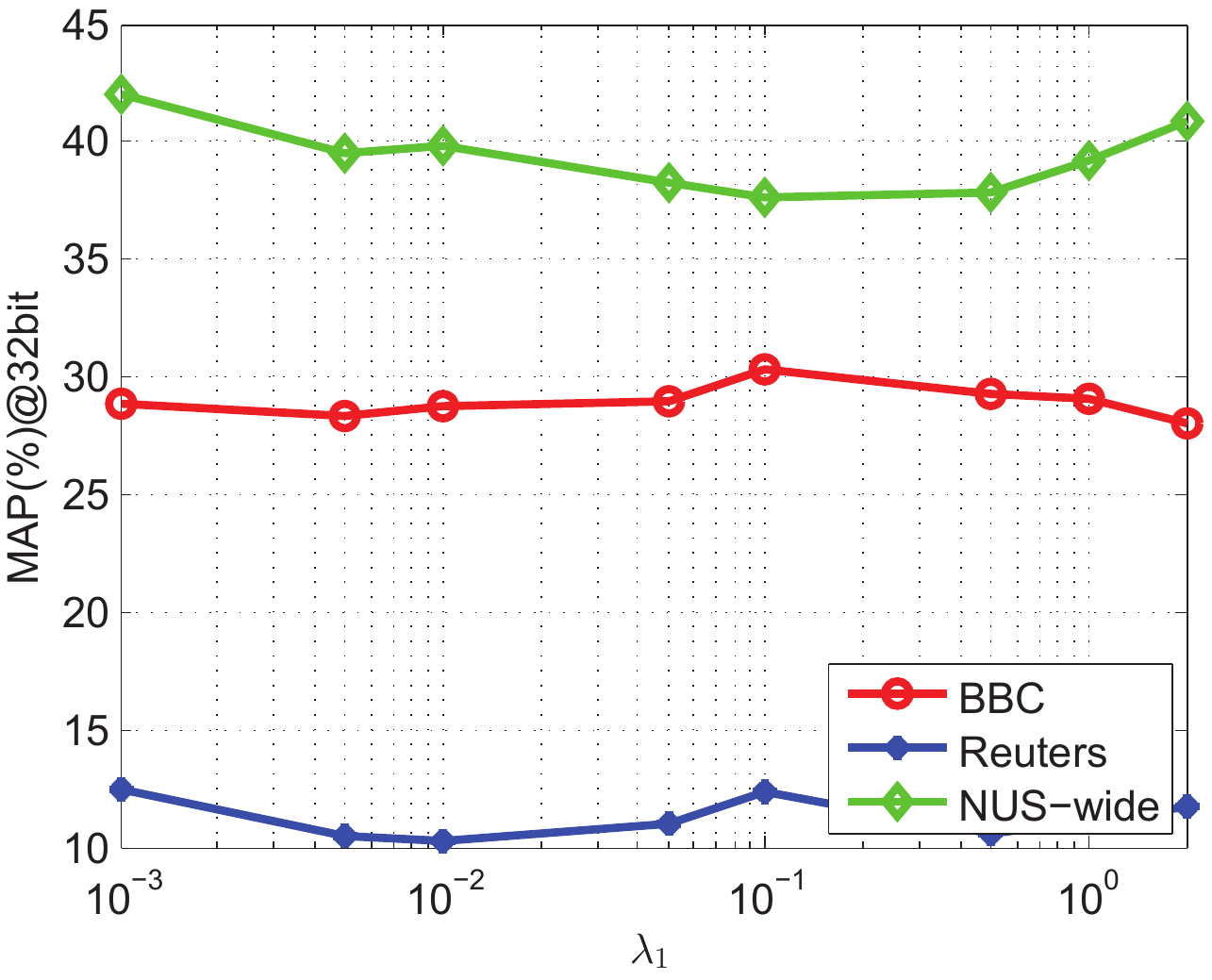}}\hspace{5mm}
\subfigure[$\lambda_2$] {\label{fig:lam2}\includegraphics[width=0.45\columnwidth]{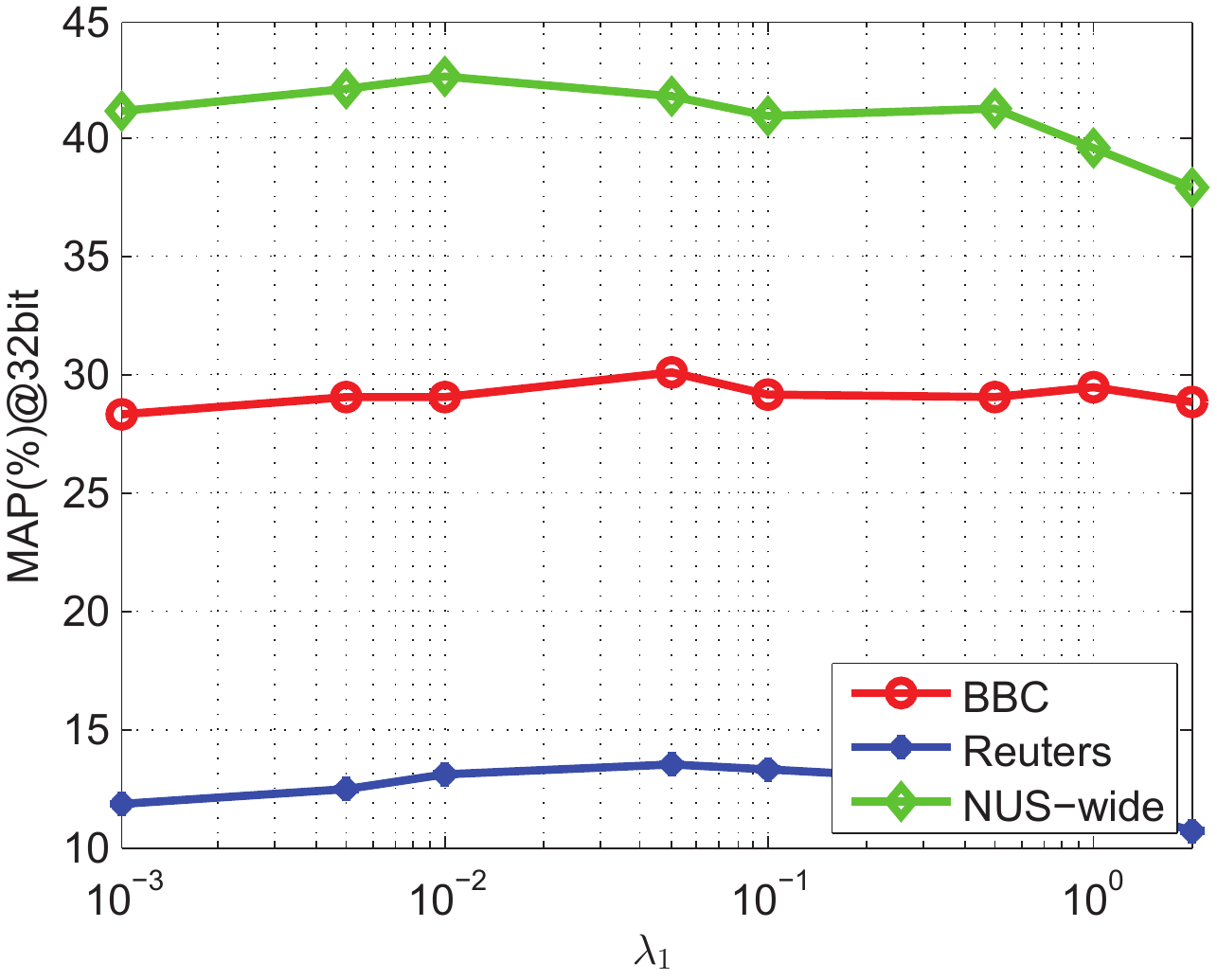}}
\caption{Parameter Analysis.}\label{fig:param}
\end{figure}
\subsection{Parameter Analysis}
In ITQ+, there is one parameter $\lambda_1$, and in LapITQ+, there are two parameters $\lambda_1$  and $\lambda_2$. As LapITQ+ is an extension of ITQ+, we only analyze the parameter sensitivity of LapITQ+ in the range of $[0,0.001,0.005,0.01,0.05, 0.1,0.5, 1, 2]$. We first fix $\lambda_2 = 0.01$ and vary $\lambda_1$. The results of LapITQ+ with 32 bits on the three datasets are shown in the Figure \ref{fig:lam1} with x-log scale. In the second experiment, we fix $\lambda_1 = 0.01$ and vary $\lambda_2$. The results are reported in the Figure \ref{fig:lam2}.\footnote{The results on the three datasets are 28.49,  11.52, 41.78 with $\lambda_1 = 0$, and 27.56, 13.37, 42.41 with $\lambda_2 = 0$, respectively.} We observe that LapITQ+ is not sensitive to $\lambda_1$ and $\lambda_2$.

\section{Conclusion}
In this paper, we propose a new learning framework for hashing named Transfer Hashing with Privileged Information (THPI), where privileged information is used to approximate a slack function to regularize the learning of hashing functions with insufficient data instances in the target domain. Based on the framework, we develop two particular transfer learning methods named ITQ+ and LapITQ+. We conduct extensive experiments on three benchmark datasets. Experimental results verify the superiority of the proposed methods ITQ+ and LapITQ+. %In our future work, we aim incorporate label or side information into our proposed framework to enable more effective semantic search.

\section{Acknowledgement}
Sinno Jialin Pan is supported by the NTU Singapore Nanyang Assistant Professorship (NAP) grant M4081532.020.
Ivor W. Tsang is grateful for the support from the ARC Future Fellowship FT130100746 and ARC grant LP150100671.
\bibliographystyle{named}
\bibliography{reference}

\end{document}